# An adaptive artificial neural network-based generative design method for layout designs

Chao Qian, Renkai Tan, Wenjing Ye*

Department of Mechanical and Aerospace Engineering

The Hong Kong University of Science and Technology, Clear Water Bay, Kowloon, Hong Kong

*Corresponding author, mewye@ust.hk

## Abstract

Layout designs are encountered in a variety of fields. For problems with many design degrees of freedom, efficiency of design methods becomes a major concern. In recent years, machine learning methods such as artificial neural networks have been used increasingly to speed up the design process. A main issue of many such approaches is the need for a large corpus of training data that are generated using high-dimensional simulations. The high computational cost associated with training data generation largely diminishes the efficiency gained by using machine learning methods. In this work, an adaptive artificial neural network-based generative design approach is proposed and developed. This method uses a generative adversarial network to generate design candidates and thus the number of design variables is greatly reduced. To speed up the evaluation of the objective function, a convolutional neural network is constructed as the surrogate model for function evaluation. The inverse design is carried out using the genetic algorithm in conjunction with two neural networks. A novel adaptive learning and optimization strategy is proposed, which allows the design space to be effectively explored for the search for optimal solutions. As such the number of training data needed is greatly reduced. The performance of the proposed design method is demonstrated on two heat source layout design problems. In both problems, optimal designs have been obtained. Compared with several existing approaches, the proposed approach has the best performance in terms of accuracy and efficiency.

## 1. Introduction

Layout design can be broadly defined as the placement of design components/materials in a domain to achieve certain design objectives. This type of problem is frequently encountered in a variety of fields. For example, the V-shaped formation of birds can be regarded as an optimized solution of a layout design problem to improve aerodynamic efficiency [1,2]. In engineering fields, the design of novel composite materials by strategically distributing soft materials inside a hard host material to achieve high stiffness and toughness [3] and the arrangement of heat sources in a system to reduce the maximum temperature [4] are all layout design problems. Often these design problems involve a large number of design degrees of freedom and/or are subjected to various constraints, making the search for the optimal solution extremely challenging and slow.

In recent years, machine learning methods have been increasingly applied in solving design problems. There are mainly two groups of machine learning-based approaches for layout designs. One group uses artificial neural networks as efficient surrogate models to speed up design evaluation in each iteration, and thus shortens the design process. For example, a feature pyramid network was constructed to rapidly predict the temperature field of a design layout directly from its structure [4]. It was then combined with the neighborhood search-based optimization procedure to find an optimized placement of heat sources to minimize the maximum temperature. In [5], a neural network-based surrogate model was developed to predict the drag force of an underwater vehicle fleet and used in the design of a fleet with minimum drag force. One issue in this type of the approach is the high computational cost associated with training data generation, which is often conducted using high-dimensional simulations, for example, the finite element analysis. The amount of training data should be sufficient so that the prediction accuracy of a surrogate model is high enough to yield satisfactory design solutions. A simple approach is to use brute force algorithm to construct a large amount of data covering the entire design space [3,6]. Such an approach is not only inefficient in problems with many design variables, but also not suitable for problems in which good design solutions lie in an extremely small portion of the design space. How to efficiently generate training data to truly achieve time saving in design problems is an ongoing research topic [7]. In a recent work [8], a machine learning framework was developed for topology optimization, in which a deep neural network (DNN) was used to predict the sensitivity information required to update the design. The online training and updating scheme and the two-scale topology optimization formulation facilitate an efficiency approach that shows promising potential to greatly reduce the computational time for large-scale topology designs.

Another group of approaches use various artificial neural networks to directly produce design solutions. For example, convolutional neural networks(CNNs) were used to map structures directly to their responses and the inverse design was carried out using a gradient-based optimization scheme in which the gradients of the responses with respect to each design variable, that is, pixel value, were obtained by performing the backpropagation of the network [9,10]. One challenge in this type of approaches is that for problems with a large number of design variables, the design converges very slowly and the computational cost increases greatly [11,12]. In addition, the obtained optimized structure likely contains unmanufacturable features. With the advent of image generative neural networks such as generative adversarial neural network (GAN) and variational autoencoder (VAE), generative machine learning methods are gaining popularity in design. In [13], a design method based on VAE model was developed. The VAE model served two functions. One was to generate design candidates and the other was to correlate the layout with the response. In a later paper [14], the two functions were separated to improve the performance. A GAN was used to generate design candidates and a CNN was used to map the design to its response/functionality. A design neural network was then constructed by connecting the generator of the GAN with the CNN to produce optimal designs. A similar approach was used to design phononic crystals with a prescribed band gap [15]. In this work, an autoencoder was used to generate crystal structures and a multi-layer perceptron was used to map a phononic structure with its band gap vector. Compared with conventional design methods such as topology optimization, the advantages of the neural network-based generative approaches are that (1) the design variables are latent variables instead of pixel values, which greatly reduce the size of the design problem, and (2) the shapes of designs can be controlled and thus the produced optimized designs are more practical. In a recent work [16], a VAE combined with a regression network was used to map complex microstructures into a low-dimensional, continuous, and organized latent space. This organized latent space facilitates the multiscale metamaterial systems design in which a diverse set of microstructures with specified properties can be rapidly generated and assembled by an efficient graph-based optimization method. However, one main issue in generative machine learning methods is that the design solutions are inherently confined within the design space generated by generative models. To obtain the desired solution, for example, the global optimal design, the generated design space must contain such a design. This is generally difficult to achieve unless a large corpus of training data that covers the entire design space is used to train the generative and the corresponding regression models, which is of course computationally intensive.

In this paper, we present an adaptive learning strategy for both generative and regression models so that a sequence of design spaces are generated. Each of these design spaces occupies a small portion of the entire design space, and hence only a small number of training data is required. Through an iterative learning and optimization process, the sequence approaches to the region that contains optimal designs, allowing the desired solution to be found. To demonstrate the performance of the proposed approach, two types of heat source layout design problems were solved using the proposed method. Such problems are frequently encountered in the thermal management of electronic devices/systems [17-20] and satellite systems [21,22]. Due to the large design space, efficient design methods are in great demand.

The remaining part of the paper is organized as follows. Section 2 introduces the heat source layout design problem and the existing methods. In Section 3, the proposed adaptive deep neural network-based design method is described. Methods for constructing GAN to produce design candidates satisfying prescribed constraints and CNN for rapid prediction of temperature field are also described. Results are presented in Section 4 including the generation performance of the constructed GAN, the prediction performance of the CNN, and design results of two heat source layout design problems. Performance comparison with conventional methods is also presented. A summary of the work is given in the last section.

## 2. Heat source layout design

The size of electronic components continuously shrinks thanks to the advent in microfabrication technology. An adverse effect of this movement is the continuously increased power density due to the densely packed electronic components. Heat concentration and hot spots are more frequently occur in these devices/systems, which undermine the operation consistency and reduce the lifespan. Thermal management has become a pressing issue in the development of this field [17,23]. Various approaches have been proposed. Among them, the method of using smart layout designs to reduce heat concentration and eliminate hot spots is perhaps the most cost-effective and easy approach to implement in practical applications, for example, to optimize the distribution of the high thermal conductivity material in the domain [24-26], or to intelligently place heat sources in the system to reduce the heat concentration and/or eliminate hot spots [27,28].

In a typical heat source layout design problem, a number of non-overlapping heat sources, for example, electronic components, are to be placed at certain locations in a domain so that the corresponding temperature field of the system satisfies a set of prescribed conditions, for example, having the lowest maximum temperature. Such a design problem can be casted into an optimization problem with the design objectives being the prescribed conditions on the temperature field and the design variables being the possible locations that each heat source can occupy. Assuming that the temperature of the system is pre-dominated by thermal conduction and/or convection and using a 2D problem as an illustration example, the governing equation for the steady state heat conduction is:

$$\frac{\partial}{\partial x}\left(k\frac{\partial T}{\partial x}\right) + \frac{\partial}{\partial y}\left(k\frac{\partial T}{\partial y}\right) + \emptyset(x,y) = 0 \quad (1)$$

where $T$ is the temperature, $k$ is the thermal conductivity and $\emptyset(x,y)$ is the intensity distribution of heat sources, which is defined by the locations of heat sources as:

$$\emptyset(x,y) = \begin{cases} \emptyset_i, & (x,y) \in \Gamma \\ 0, & (x,y) \notin \Gamma \end{cases} \quad (2)$$

where $\emptyset_i$ is the intensity of the $i$-th heat source ($i = 1, 2 \ldots n$) and $n$ is the totally number of heat sources; $\Gamma = \bigcup_{i=1}^{n} \Gamma_i$ denotes the region occupied by heat sources and $\Gamma_i$ denotes the region occupied by the $i$-th heat sources.

Different boundary conditions can be applied depending on applications. Three common boundary conditions are the adiabatic boundary, the isothermal boundary and the convective boundary, which can be formulated as follows:

$$T = T_0, \quad k\frac{\partial T}{\partial n} = 0, \quad k\frac{\partial T}{\partial n} = h(T - T_1) \quad (3)$$

where $T_0$ represents the temperature of the isothermal boundary; $h$ is the convective heat transfer coefficient and $T_1$ is the fluid temperature of the convective boundary.

Due to the large number of possible locations that heat sources can occupy, efficient design methods must be developed. Over years, several methods have been proposed such as the bionic optimization method [27], the gradient-aware bionic optimization method [29], the simulated annealing method [28] and iterative reweighted L1-norm convex minimization [30]. The bionic optimization method [27] aims at achieving a uniform temperature field in the design domain to eliminate hot spots within the domain. In this method, heat sources are added one by one and always at the location with the minimum temperature. The gradient-aware bionic optimization method [29] is an improved version of the bionic optimization method. It aims at achieving a uniform temperature gradient instead of temperature to eliminate hot spots. When a new heat source is introduced, the gradient-aware bionic optimization method places it at the location with the maximum temperature gradient so that a uniform temperature gradient is achieved in the final design. Both methods are efficient, but they are not general. For example, the temperature constrained heat source layout design problem [4] will be difficult if not impossible to be solved by these two methods. The simulated annealing method reduces the maximum temperature by placing heat sources through random walks based on Metropolis principle [28]. The convergence of this method is very slow, and consequently the corresponding computational cost is very high. In [30], the heat source layout design problem was transferred into a convex optimization problem and the sequential reweighted L1-norm minimization technique was used to solve the problem. An advantage of this method is that only one full-scale temperature simulation is needed, and thus the computational cost is greatly reduced. However, the design results produced were not as good as those obtained from other methods. In addition, a nonzero inter-source spacing must be set to avoid the overlapping of heat sources, which introduces additional errors.

Apart from these conventional approaches, machine learning methods have also been applied to solve heat source layout designs. In [31], a layout design with three heat sources was performed in a 2D ventilated cavity cooled by forced convection. An artificial neural network was constructed to predict the maximum temperature of the cavity. It was then integrated with the genetic algorithm (GA) to optimize the locations of the three heat sources to reduce the maximum temperature in the ventilated cavity. A similar approach was employed in the layout design of five heat sources in a three-dimensional vertical duct [32]. In this case, due to the small size of the design space, an exhaustive search approach was used to generate all possible configurations. In [4], a feature pyramid network was firstly trained to predict the temperature field of heat source layout designs and then combined with the neighborhood search-based optimization to search for the optimal heat source layout with the lowest the maximum temperature. In all the work, an artificial neural network is used to speed up the calculation of the maximum temperature. However, the efficiency gained by using artificial neural network is largely reduced by the large amount of training data used to train the artificial neural network.

3. Design methods

3.1 Brief overview of the layout design method based on GAN and CNN

The framework of the GAN+CNN-based design method proposed in [14] consists of three parts: a design generator, a design evaluator and a design network as illustrated in Fig. 1. Design candidates are generated by the design generator, which is a generative neural network such as a GAN. There are two advantages of using a generative neural network. First, the layout/structure is described by the latent vector of the generative model, whose dimension is much smaller than the number of design variables. Secondly, geometric constraints of the design can be enforced through the training of the generative model. For example, to design composite materials with only circular inclusions, the GAN can be trained with various microstructures with circular inclusions. Once trained, the generator of the GAN will generate new microstructures with circular inclusions. The design evaluator, which is typically a convolutional neural network, outputs the performance of a given candidate design, for example, the maximum temperature. Then the generator of the GAN and the CNN are connected to form the design network on which the inverse design is conducted. The design variables are the latent variables, $Z$, of the generative network, and hence the number of design variables is greatly reduced.

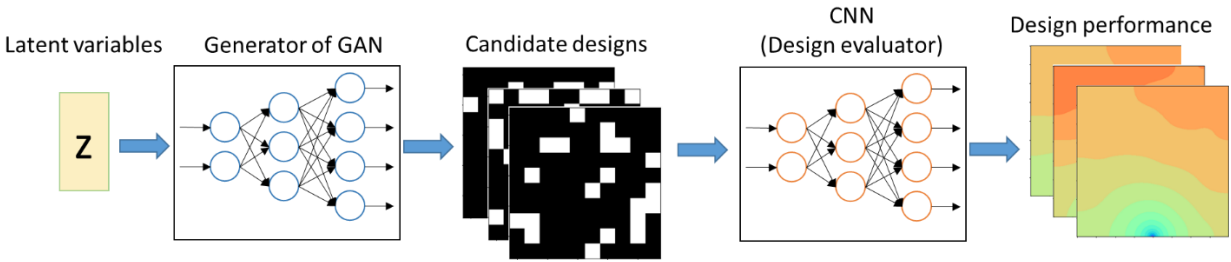

Fig. 1. Schematic of the framework of the GAN+CNN-based design method.

For a given design target, an optimization scheme is employed to search for a particular $Z$, which produces a design with the target performance. In [14], a gradient-based optimization scheme was used. The gradient of the loss function with respect to the design variable, that is, $Z$, was calculated via backpropagation of the design network. Gradient-based optimization schemes are efficient, but the design solution produced highly depends on the initial guess. Hence in [14], several runs of optimization corresponding to different initial $Z$ were conducted and the best design solution was selected among the design solutions obtained from these runs.

3.2 The adaptive learning and optimization design method based on GAN, CNN and genetic algorithm

In the original GAN+CNN design method, the two networks, that is, GAN and CNN, are separately trained and conducted off-line. Once trained, these two networks are then combined to form the design network. The training data for both networks must be sufficiently large to ensure that the design space covered by the GAN is large enough to contain the desired design solution. Hence, a large amount of training data is often required. Although the training for GAN is unsupervised, the regression model, CNN, needs to be trained with labels, and often high-dimensional simulations are used to produce these labels. Therefore, the computational cost is significant and can largely diminish the efficiency gained by using neural networks.

In this work, an adaptive learning and optimization strategy was proposed and employed to train the GAN and the CNN. The key idea of this strategy is to start with a small design space and then by using iteratively training and optimization gradually move the design space towards to the region that contains the desired design solution. To do so, the training of the GAN and the CNN is conducted online and guided by the optimization results via an iterative procedure. The training data for the GAN changes at each iteration, but the total number is fixed so the design space covered by the GAN remains small. The training data for the CNN is a small subset of the training data for the GAN. To ensure prediction accuracy, the training data for the CNN is continuously expanded by a fixed number at each iteration. The flowchart of the proposed design approach is shown in Fig. 2. We first generate $M_0$ random layouts and randomly selected $N_0$ layouts with $M_0 \gg N_0$ to serve as the training data for the CNN. The Finite Element Method (FEM) or any other method is used to generate labels for these layouts, for an example, to calculate the temperature field of each layout in the heat source layout design problems. The trained CNN is used to screen the original $M_0$ layouts and $M_1$ layouts with best performance are selected. Next, we perform layout perturbation on each selected layout by randomly selecting a heat source and moving it to one of its immediate neighboring pixels, that is, either the left or the right or the top or the bottom pixel of the heat source. This process is conducted *p* times sequentially, resulting in *p* new layouts for each selected layout. This process is necessary to move the design space towards the region that contains better solutions. All together they form $(p+1)M_1$ layouts. The CNN is then used to screen these layouts and $M_1$ layouts with the best performance are selected to serve as the initial training data for the GAN. Among $M_1$ layouts, we randomly select *K* images and use the FEM to generate their labels. These images together with their labels are added to the original $N_0$ training data to form a new set of training data for the CNN with the total number of $N_1 = N_0 + K$. After the CNN is re-trained using this new set of data, the generator of the GAN and the CNN are connected to form the design network and the inverse design is performed using this network in conjunction with the genetic algorithm (GA). The design variables are the latent variables of the GAN and the design performance is evaluated using the design network. GA is a gradient-free optimization algorithm and has the potential to produce the global optimal solution [33-35]. Another reason for choosing GA instead of a gradient-based optimization scheme in this work is that in layout design problems, the objective function is often non-differentiable with respect to design variables.

Since a small set of training data is used to train the GAN and the CNN, the design solutions obtained from the GA are generally not satisfactory since the design space is confined within the space covered by the GAN. To improve results, more design iterations need to be conducted. For each design in the final generation of GA, layout perturbation is performed to generate new designs. Among these new designs and the designs in the final generation of GA, the best $M_1$ designs are selected using the CNN trained in the last design iteration as the screening tool. These $M_1$ designs are used to re-train the GAN. In the meantime, the training data of the CNN is expanded by adding *K* new data randomly selected from $M_1$ designs and the CNN is re-trained as well. The re-trained GAN and CNN form a new design network and the inverse design is performed again. This process repeats until the best design obtained converges.

Unlike the previous GAN+CNN-based design method, the proposed method trains networks and conducts the design iteratively. In each iteration, the design solutions obtained from the GAN+CNN+GA method at the previous iteration are collected. They are expanded by layout perturbation and then used to re-train the GAN. Thus, the design space covered by the GAN changes at each iteration and gradually approaches the region containing optimal solutions even though the initial space covered by the GAN does not contain the desired solutions. The training data for the CNN is gradually expanded at each iteration with newly added data being a subset of the training data for the GAN at each iteration. Hence the prediction accuracy of the CNN improves with the number of iterations. This process allows us to effectively use training data and greatly reduce the computational cost. Such a process is most effective in cases with no physical intuition on good designs.

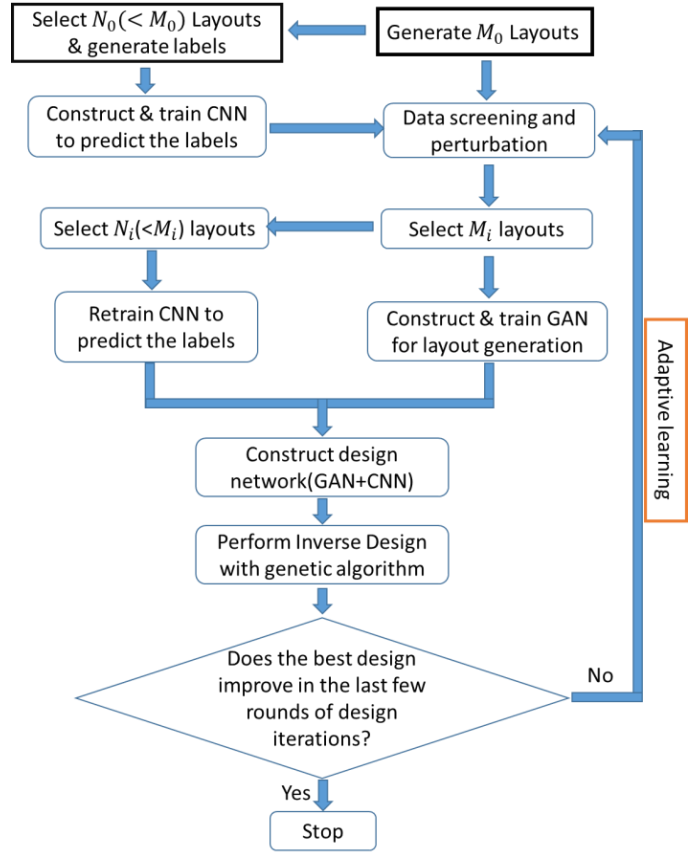

Fig. 2. The flowchart of the adaptive learning and optimization approach for layout design ($i$ denotes the $i$-th round of inverse design).

3.3 Generative adversarial network for heat source layout generation

GAN is a type of generative neural network [36], which has been widely applied in various applications such as generating porous material structures [14], handwritten digits [37] and anime characters [38]. A GAN consists of a generator and a discriminator. The generator of the GAN produces images while the discriminator determines whether the image is real or fake. Through an adversarial training process, the generator gradually learns the distribution of the training data and can generate images with the same characteristics of the training data.

In this work, GAN is used to generate candidate layouts of heat sources. A major challenge in this task is the precise control of the number of the heat sources. Although GAN has been used to produce different images including structures and metamaterials [39-42], none of them need to have the precise control on certain image properties. One approach to control the properties of images is to use a conditional GAN [43]. However, we find that it is rather difficult to keep the number of heat sources exactly at the prescribed value in all images generated by the conditional GAN. We then develop a simple method to precisely control the number. Fig. 3 shows the architecture of the GAN used in this work. It contains one fully connect layer, three convolutional layers, followed by a controlling layer and the final layer. Except for the controlling layer, all other layers are common layers in a conventional GAN. The values in the controlling layer are obtained by subtracting a controlling parameter, $VF$, from values in the previous layer, that is, $X_i^{N-1} = X_i^{N-2} - VF$. Through sigmoid projection and rounding, the values in the final layer, that is, $X_i^N = [sigmoid(X_i^{N-1})]$, where [·] represents rounding, are either 1 or 0 with "1" indicating the existence of a heat source at that element and "0" representing the background material. The value of the controlling parameter, $VF$, is chosen according to the required number of heat sources. For example, if five heat sources are required, $VF$ takes the average value of the fifth and sixth largest elements in the previous layer, so that there will be exactly five positive elements after the subtraction and thus five heat sources in the final image.

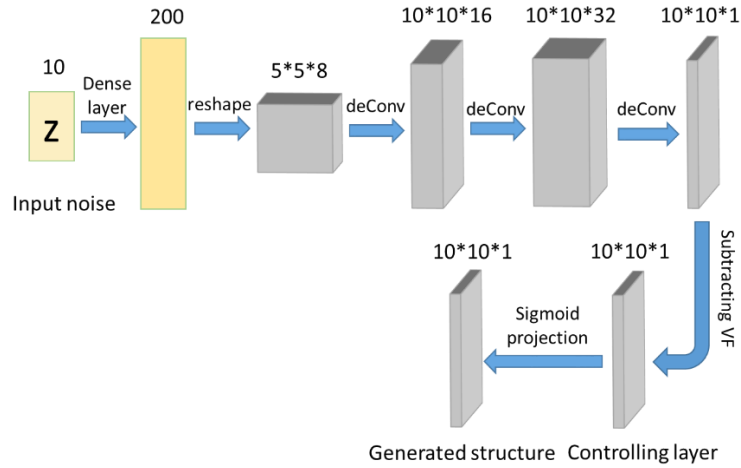

Fig. 3. The architecture of the generator of the GAN with a controlling layer to control the number of the heat sources.

*3.4 ThermalNet*

Artificial neural networks have been used increasingly as surrogate models for rapid predictions of physical quantities/fields. Once constructed, artificial neural networks can perform the prediction often in milliseconds as opposed to minutes or hours that high-dimensional physics-based simulations need. A key step in design problems is the repeated evaluations of the objective function. Hence in this work, a convolutional neural network named ThermalNet is constructed to predict the temperature field of the designs. The architecture of the ThermalNet is shown in Fig. 4, which contains a few convolutional and deconvolutional layers and Se-resNet layers[44].

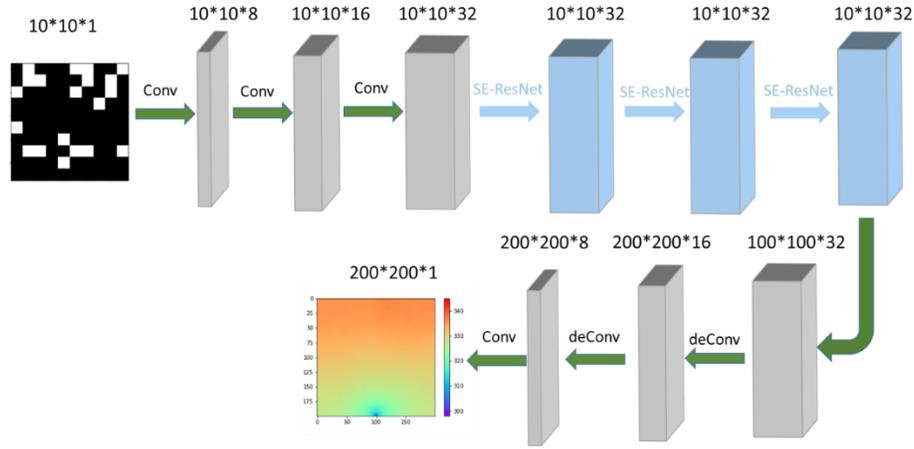

Fig. 4. The architecture of ThermalNet for temperature field prediction.

The input of the ThermalNet is an image of the heat source layout, which is represented by a 10 by 10 binary matrix. The output is the corresponding temperature field. To accurately capture the temperature field, the resolution of the output temperature field is set to be 200 by 200, much finer than that of the layout image.

## 4. Results

### 4.1 GAN's performance

The function of the GAN in this work is to generate layouts with 20 square-shaped heat sources. To do so, images with 20 non-overlapped heat sources are generated using random sampling and the evolving sampling method proposed in [4]. Evolving sampling allows images with "extreme" distribution of heat sources to be produced, for example, all heat sources located at the top two rows of the image. Fig. 5 shows some of these images where white elements denote heat sources and black elements denote the background material. To test the effect of the controlling layer for the number of heat sources, ten thousand of these images are used to train the GAN with and without the controlling layer. After training, 2000 images are generated by the each GAN. Some generated images are shown in Fig. 6. In all the 2000 generated images, heat sources remain square shaped and are non-overlapped. As expected, the number of heat sources in images generated by the GAN without the controlling layer varies, ranging from 10 to 30 as shown in Fig. 6(a). On the other hand, those generated from the GAN with the controlling layer, which are shown in Fig. 6(b), have the exact 20 heat sources in each image, indicating the effectiveness of the controlling method.

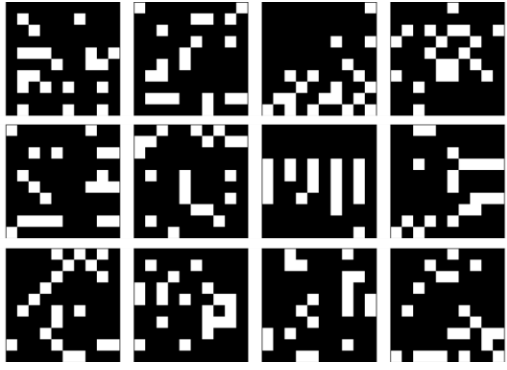

Fig. 5. Some training samples of the GAN.

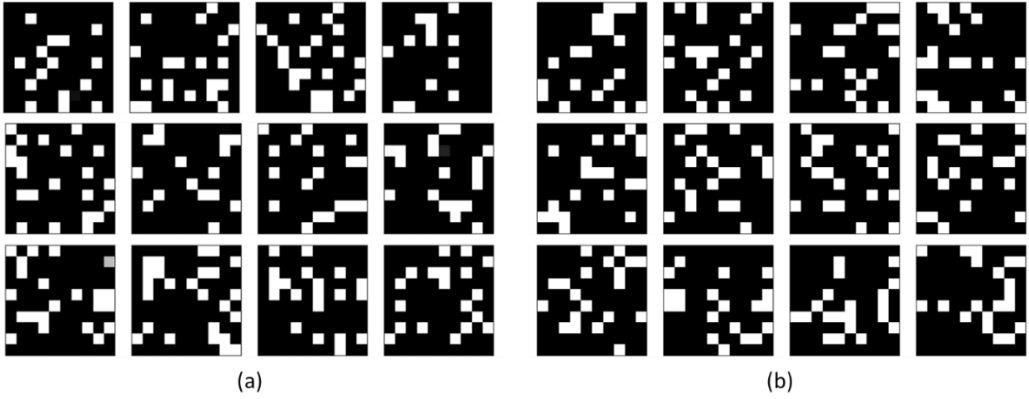

(a)  (b)

Fig. 6. Samples of the generated heat source layouts of the GAN: (a) without the controlling layer; (b) with the controlling layer

4.2 ThermalNet's performance

To domonstrate the performance of the ThermalNet for temperature prediction, 10000 images are collected as the training data. Among them, 6000 are generated by sampling and 4000 are generated by evolving sampling. Their temperature fields are calculated using the FEM and serve as the labels. To test the prediction accuracy of the ThermalNet, additional 50 layouts and their corresponding temperature fields are generated using the same approach. The testing performance of the trained ThermalNet is shown in Fig. 7, where the normalized temperature of each pixel/element in each image is plotted with the horizontal axis being the groundtruth and the vertical axis being the predicted values from the ThermalNet. Here the normalization of the temperature field is defined as $\frac{T-298K}{350K-298K}$, so it is within the normal range of neural network's ouput, that is, -1 to 1. The red line with unit slope indicates the case when the prediction perfectly matches with the groundtruth and the two dash lines indicate the $\pm 5\%$ error range. As evident from the figure, all dots are within the two dash lines with most of them located very close to the red line. Fig. 8 shows the temperature fields of two testing samples. Overall the predicted temperature fields shown in the right column agree well with those shown in the middle column, which are calculated from the FEM. The average relative error of the temperature field defined as $\frac{1}{N}\sum_{i=1}^{N}\frac{|T_i-\tilde{T}_i|}{T_i}$, where $T_i$ and $\tilde{T}_i$ are the predicted temperatrue and the goundtruth at the $i$-th pixel respectively, is 1.41%, indicating the high prediction accuracy of the ThermalNet. Regarding the efficiency, the FEM takes 5.4s to compute the temperature field of one image on a computer with 2 x Intel Xeon E5-2670 v3 (2.3GHz), while the ThermalNet uses only 0.0018s to generate the temperature field per image on the same machine. An improvement of three orders of magnitude is achieved.

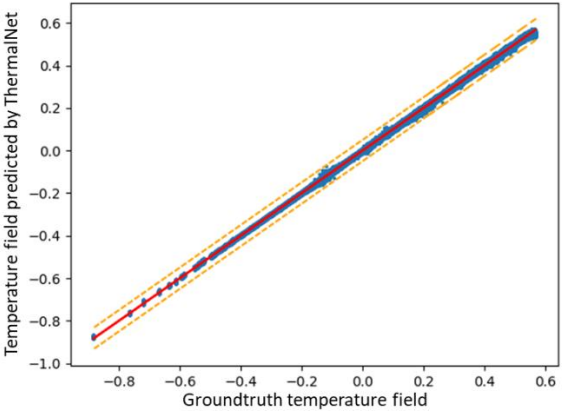

Fig. 7. The prediction accuracy of the trained ThermalNet. Each dot represents the temperature of a pixel in a testing image. A total of 2,000,000 dots are plotted. The red line with unit slope indicates the case when the prediction perfectly matches with the groundtruth and the two dash lines indicate the $\pm 5\%$ error range.

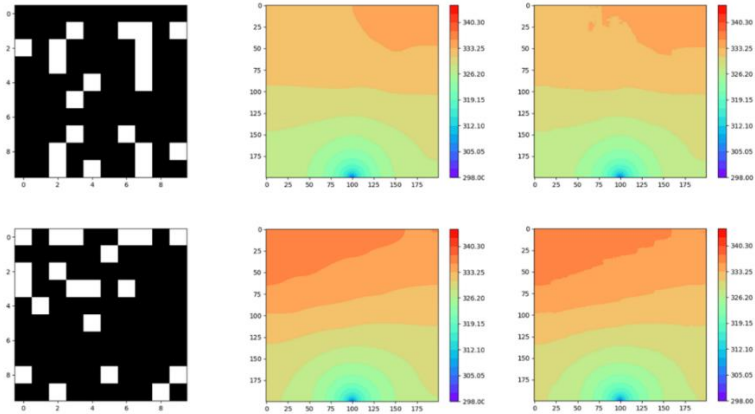

Fig. 8. Temperature fields of two testing samples: the layouts (left); temperature fields predicted by the FEM (middle); temperature fields predicted by the ThermalNet (right).

The good accuracy shown above is achieved using 10k training data and is also because that the testing samples lie in the space covered by the training data. If layouts are quite different from those images in the training set, the prediction accuracy will deteriorate. In some design problems, optimal layouts have rather extreme distributions that are difficult to generate by the random sampling approach and even the evolving sampling method. The design of heat source layout with minimum maximum temperature is one of such cases. Based on the physical knowledge, designs with small maximum temperature should be those with all heat sources located near the heat sink. One hundred of such layouts are generated manually and their temperature fields are computed using the ThermalNet. Fig. 9 shows the prediction accuracy of each image in the 100 layouts as compared with the groundtruth values. Most data points are outside the $\pm 5\%$ error lines and the average relative error of all data points is 27.3%, much higher than 1.4% achieved when the testing samples are within the space covered by the training data, indicating that even with 10000 samples generated via random and evolving sampling, the space covered by the training data was not large enough to contain these extreme layouts. Fig. 10 shows the temperature fields of two layouts that have low maximum temperature. The temperature fields produced by the FEM (middle column) and by the ThermalNet (right column) are quite different. The relative errors of the maximum tempertaure of the two layouts predicted by the ThermalNet are 120.56% and 115.87% respectively, demonstrating that the ThermalNet trained using 10k data is not adequete for this design problem. A much larger training data set would be required if the same data generation method were used. This would greatly increase the computational cost and diminish the efficiency gained by using the ThermalNet. In the next section, we will show that by using the adaptive learning strategy, a much smaller training set is sufficient to produce good design solutions.

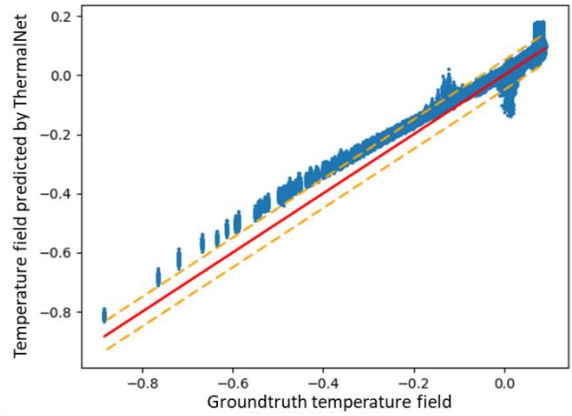

Fig. 9. The prediction accuracy of the trained ThermalNet on 100 "extreme" samples. Each dot represents the temperature of a pixel in a testing sample. The red line with unit slope indicates the case when the prediction perfectly matches with the groundtruth and the two dash lines indicate the $\pm 5\%$ error range.

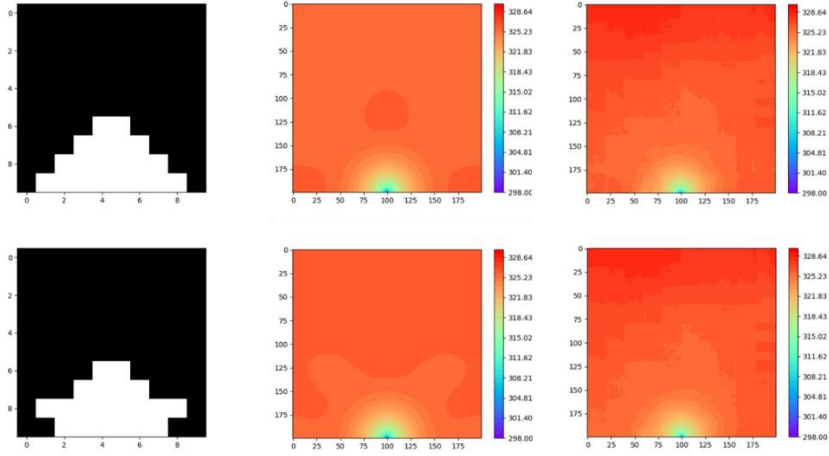

Fig. 10. Temperature fields of two layouts with low maximum temperature: the layouts (left); temperature fields predicted by the FEM (middle); temperature fields predicted by the ThermalNet (right).

4.3 Heat source layout design with minimum maximum temperature

The first design problem solved using the adaptive learning and optimization method is the layout design with minimum maximum temperature. Fig. 11 shows the schematic of the problem domain, which is a square domain with length $L = 0.1m$. Twenty square heat sources with length of $L = 0.01m$ are to be distributed within this domain and the design objective is to minimize the maximum temperature of the domain. All four sides of the material domain are adiabatic except the middle region of the bottom boundary where its temperature is kept at a constant temperature $T_0 = 298K$. Hence this region serves as the heat sink for heat flowing out of the domain. The length of the heat sink is $\delta = 0.001m$. The thermal conductivity of the background material is $k = 1W/(m \cdot K)$ and the intensity of each heat source is $\emptyset_0 = 10000W/m^2$. The design domain is discretized into a uniform 10 by 10 grid and each heat source occupies one element/pixel as shown in Fig. 12. For the FEM simulation of the temperature field, the design mesh is further refined into a uniform FEM mesh with 200 by 200 elements. This is to ensure the accuracy of the simulated temperature field.

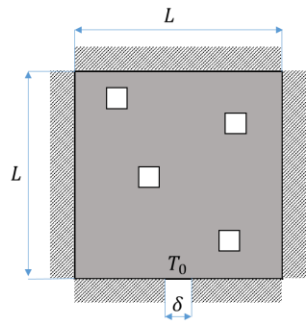

Fig. 11. Schematic of the problem domain of the heat source layout design with minimum maximum temperature.

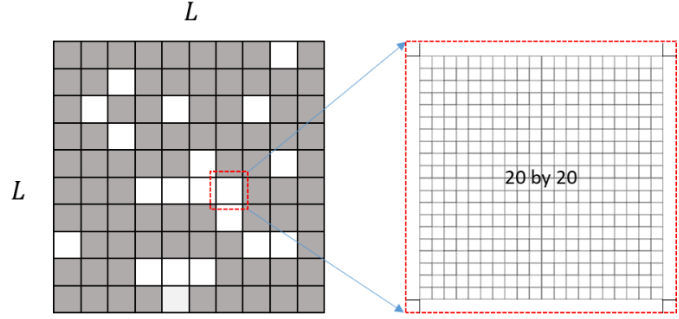

Fig. 12. The design mesh for the heat source placement and the mesh for FEM simulations: the white region indicates the heat source and the gray region represents the background material.

Genetic algorithm is employed to solve the inverse design problem. The population size is set to be 250 and a relatively large mutation rate 20% is used considering that this is an extreme design problem. The number of design variables, that is, the latent variables of the GAN, is 10, much smaller than the 100 original design variables if GAN is not used. Adam optimizer [45] is used for the training of neural networks. The corresponding learning rates are 1e-3 and 5e-4 for the ThermalNet and the GAN respectively. The batch size is 10 for the ThermalNet due to the small size of labeled training data. And the batch size for the GAN is 250.

In the first design iteration, 20k ($M_0 = 20000$) heat source layouts are generated using the random and evolving sampling method. Among them, 250 layouts ($N_0 = 250$) are randomly selected and their temperature fields are calculated using the FEM. These layouts and their temperature fields serve as the initial training data for the ThermalNet. The trained ThermalNet is then used to screen the 20k layouts and the top 2000 layouts, that is, those with low maximum temperatures, are selected. Layout perturbation is then performed on each selected layout 10 times to produce 20000 new layouts. These new layouts together with the original 2000 layouts go through another screening and the top 2000 ($M_1 = 2000$) are selected and used to train the GAN. Fig. 13 shows some of layouts in the training set for the GAN. The majority of heat sources locate near the heat sink, which is the result of the pre-screening. Among the 2000 layouts, 62 layouts are randomly selected ($K = 62$) and are added into the initial 250 layouts ($N_1 = N_0 + K = 312$) to form an expanded training set for the CNN. The re-trained CNN and the GAN are then connected to form the design net and the GA is used to obtain design solutions. The GA is terminated when the best design in the final generation does not improve further. Fig. 14 shows some of the design solutions. Compared to the layouts in the training data for the GAN (Fig. 13), heat sources are much closer to the heat sink in the obtained designs and therefore the maximum temperatures of these designs are lower than those of the training layouts.

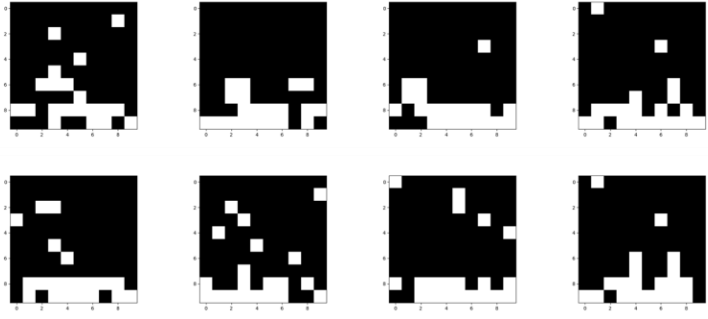

Fig. 13. Samples of the selected training data for the GAN in the first design iteration.

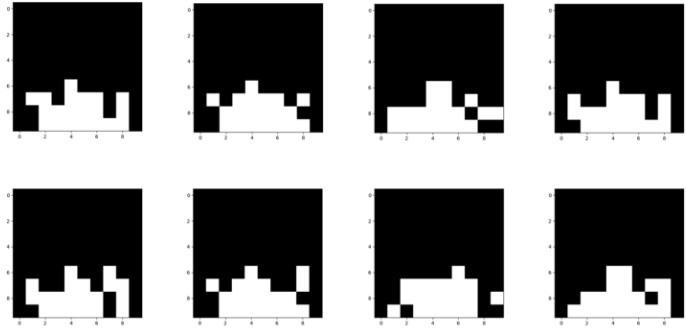

Fig. 14. Samples of the design solutions obtained from the combined network in the first design iteration.

In the subsequent design iterations, the final design solutions ($Q = 250$) obtained from the previous design iteration are used as the seeds and layout perturbation is performed 20 times on each layout. The newly generated layouts, the design solutions obtained from the previous iteration and the training layouts for the GAN in the previous iteration ($M_{i-1} = 2000$) are merged into one large data set and the CNN screening is performed on this set to select the top 2000 layouts. These 2000 layouts are used as the training data for the GAN in the current design iteration ($M_i = 2000$). For the training of the CNN, again $K$ layouts ($K = 62$) are randomly selected from the current training data for the GAN ($M_i = 2000$) and they are added to the previous training data ($N_{i-1}$) for the CNN to form the training set in the current iteration ($N_i = N_{i-1} + K$). Hence, the training data set for the CNN expands by $K$ data points each iteration. The training data for the GAN changes at each iteration but the total number is kept the same ($M_i = 2000$). To keep the training data set for the CNN small to save the computational cost, $K$ is set to be a quarter of the initial training data set, $N_0 = 250$. The trained CNN and GAN are connected to form a new design network and the inverse design is conducted using the GA. Such an iterative process continues until the design solutions obtained from the current iteration agree with those in the previous iterations within a predefined tolerance.

Table 1 shows the distribution of the maximum temperature of the training data for the GAN and that of the design solutions at the first two and the final design iterations. Principal component analysis is also performed on these layouts and the first two principal components are plotted and shown in the table. Colors in the PCA plots indicate the value of the maximum temperature of each layout. At each iteration, the maximum temperature distribution of design solutions obtained from the GA is much narrower than that of the training data and is centered at a lower temperature region. Both the average and the lowest maximum temperatures are much smaller than those of the training layouts, demonstrating the effectiveness of the inverse design. As the iteration goes on, the maximum temperature distribution of the training data continuously narrows down and moves towards the lower temperature region, indicating the design space covered by the GAN shifts towards the space containing better and better design solutions. Such a trend can also be observed in the PCA plots, where data are more clustered and have more uniform lower maximum temperature after each iteration. This shift in the design space is the main aim of the proposed adaptive learning and optimization approach, which allows us to find optimal designs without the need of a large training data. As shown in the last column of Table 1, best layout designs in selected iterations are shown and their maximum temperatures calculated from the ThermalNet and the FEM are also listed. The prediction accuracy of the ThermalNet improves with the number of iterations as expected. A total of 7 design iterations are needed in this problem. The best design obtained, which is believed to be one of the optimal solutions, has a maximum temperature of 326.06 K and it is found using a total of 622 training data for the CNN, that is, 622 FEM calculations of the temperature field.

Table 1: Results of heat source layout design with minimum maximum temperature at selected design iterations: maximum temperature distribution of data sets; principal component analysis of data sets; best designs and the corresponding maximum temperatures.

| Design iteration | | Maximum temperature distribution | Principal component analysis | Best design | |
| --- | --- | --- | --- | --- | --- |
| | | | | $T_{max}^{NN}$/K | $T_{max}^{FEM}$/K |
| 1st design iteration | Training data | 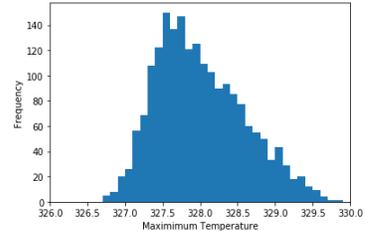 | 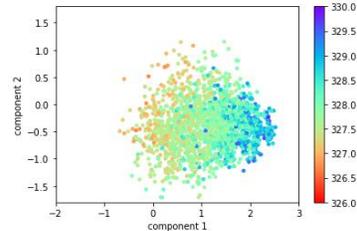 | 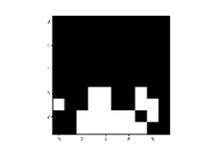 327.44 | 326.71 |
| | Generated solution | 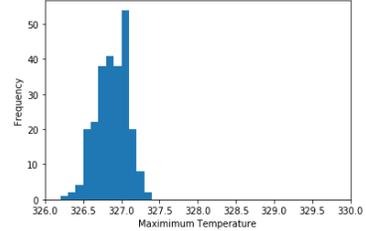 | 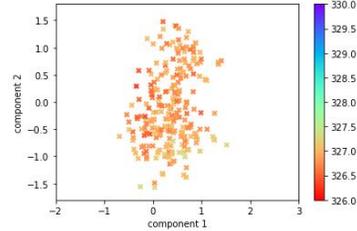 | 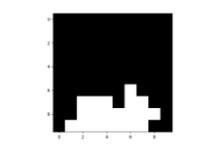 326.66 | 326.28 |
| 2nd design iteration | Training data | 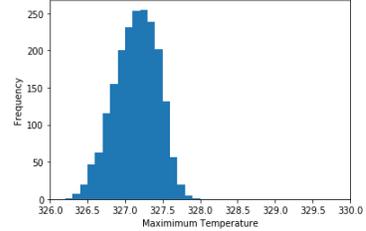 | 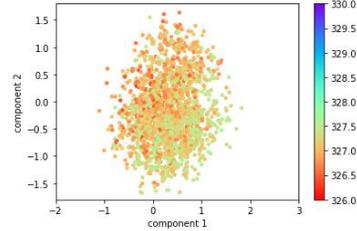 | 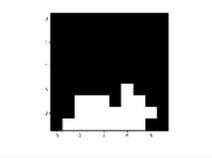 326.66 | 326.28 |

| | | | | | | |
|---|---|---|---|---|---|---|
| | Generated solution | 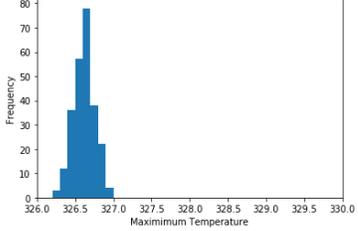 | 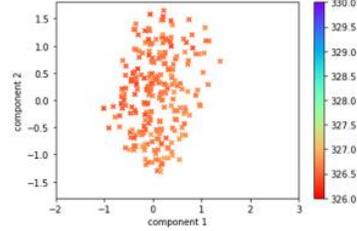 | 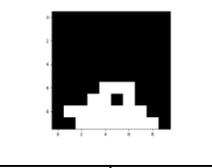 | 326.43 | 326.23 |
| Final design iteration | Training data | 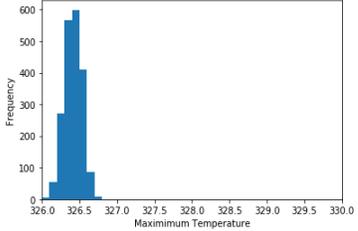 | 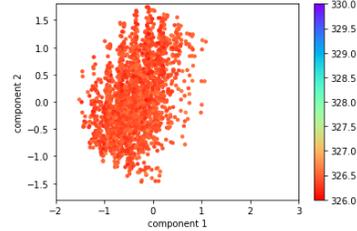 | 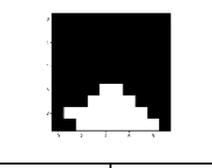 | 326.22 | 326.08 |
| | Generated solution | 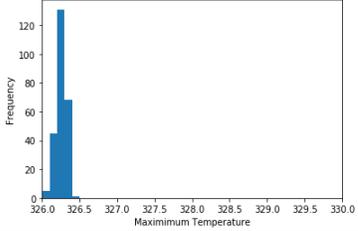 | 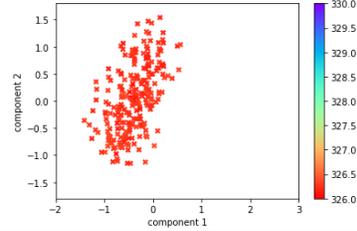 | 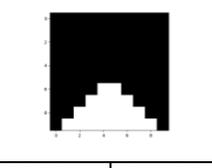 | 326.22 | 326.06 |

*Performance comparison with other design approaches*

As mentioned in Section 2, the heat source layout design with minimum maximum temperature has been solved by a number of approaches including both conventional and machine learning based methods. Table 2 summarizes the maximum temperature of the best design obtained from each of these approaches and the number of FEM simulations needed to obtain the best design. As a comparison, the maximum temperature of the best layout design obtained from the proposed adaptive GAN+CNN+GA and the number of FEM calculations are also listed. It is clear from the table that the adaptive GAN+CNN+GA produces the best design among all approaches. In addition, it is also the most efficient as compared with previous approaches except for the convex optimization method [30] in which only one FEM calculation was needed. However, the accuracy of the convex optimization method is less satisfactory. Compared with the artificial neural network-based method [4], the adaptive GAN+CNN+GA not only produces a better design but also uses a much smaller number of training data. Even with only one design iteration, the maximum temperature of the best design obtained from GAN+CNN+GA is 326.94K, which is the lowest among all other approaches. It is achieved with only 312 FEM calculations. With a few more design iterations and a total of 622 FEM calculations, a design that has a maximum temperature of 326.06K can be obtained. Hence, it can be concluded that the proposed method performs the best in terms of both accuracy and efficiency. To demonstrate the effect of the adaptive strategy, design is carried out using the proposed method without any adaptive learning. With the same number of training data for the CNN and the GAN, the best design obtained from the approach without adaptive learning has a maximum temperature of 326.94K, even higher than that of the design obtained by using only one design iteration.

Table 2: Design results obtained from different layout design methods (BO: bionic optimization [27]; TGBO: temperature-gradient aware bionic optimization [29]; SA: Simulated annealing [28]; CO: convex optimization [30]; FPN-NSLO: Feature-pyramid-network and neighborhood search-based layout design [4]).

| different layout design methods | Number of FEM calculations | $T_{max}$ of the best design/K |
|---|---|---|
| BO | 420 | 328.69 |
| TGBO | 420 | 328.70 |
| SA | 1760 | 329.61 |
| CO | 1 | 328.05 |
| FPN_NSLO | 55000 | 327.02 |
| GAN+ CNN without adaptive learning | 622 | 326.94 |
| GAN+ CNN+ adaptive learning | 312 | 326.28 |
|  | 622 | 326.06 |

*Influence of the size of ThermalNet's training data on the performance of the design method*

A large portion of the computational cost of the proposed design method is the time spent on generating the training data for the ThermalNet, which is carried out using the FEM. In this method, the size of the ThermalNet training data is determined by two parameters. They are the initial training data $N_0$ and the added training data at each iteration $K$. In this section, the influence of these two parameters on the performance of the method is investigated. Four cases with initial training data size being 1000, 500 250 and 125 are studied for this purpose. In each case, the added data is set to be a quarter of the initial data size, that is, $K = N_0/4$. Table 3 lists the number of ThermalNet training data, the best design and its maximum temperature predicted by the ThermalNet and the FEM in each design iteration for the four cases.

Table 3: The influence of ThermalNet training data on the performance of the proposed method in heat source layout design problem: $N_0$ is the initial training data size, $K = \frac{N_0}{4}$ is the added training data in each design iteration, $N_i = N_0 + i * K$ is the total number of training data in the $i$-th design iteration; $T_{max}^{NN}$ and $T_{max}^{FEM}$ are the maximum temperatures predicted by the ThermalNet and the FEM respectively.

| Design cases |  | $N_0 = 1000, K = 250$ |  |  | $N_0 = 500, K = 125$ |  |  | $N_0 = 250, K = 62$ |  |  | $N_0 = 125, K = 31$ |  |
|---|---|---|---|---|---|---|---|---|---|---|---|---|
| Design iteration | $N_i$ | Best design |  | $N_i$ | Best design |  | $N_i$ | Best design |  | $N_i$ | Best design |  |
|  |  | $T_{max}^{NN}/K$ | $T_{max}^{FEM}/K$ |  | $T_{max}^{NN}/K$ | $T_{max}^{FEM}/K$ |  | $T_{max}^{NN}/K$ | $T_{max}^{FEM}/K$ |  | $T_{max}^{NN}/K$ | $T_{max}^{FEM}/K$ |
| 1 | 1250 | 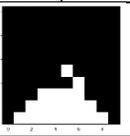 |  | 625 | 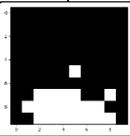 |  | 312 | 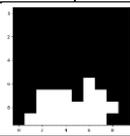 |  | 156 | 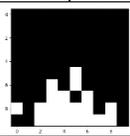 |  |
|  |  | 326.82 | 326.24 |  | 326.80 | 326.28 |  | 326.66 | 326.28 |  | 327.93 | 326.52 |

| | | | | | | | | |
|---|---|---|---|---|---|---|---|---|
| 2 | 1500 | 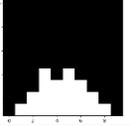<br>326.51 \| 326.14 | 750 | 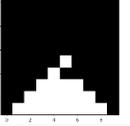<br>326.62 \| 326.20 | 374 | 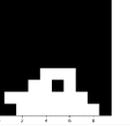<br>326.43 \| 326.23 | 187 | 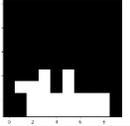<br>327.82 \| 326.48 |
| 3 | 1750 | 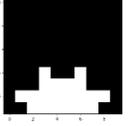<br>326.39 \| 326.06 | 875 | 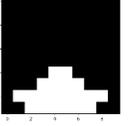<br>326.16 \| 326.09 | 436 | 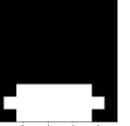<br>326.16 \| 326.15 | 218 | 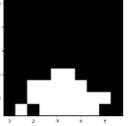<br>326.71 \| 326.18 |
| 4 | 2000 | 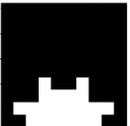<br>326.26 \| 326.06 | 1000 | 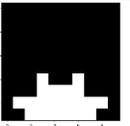<br>326.18 \| 326.06 | 498 | 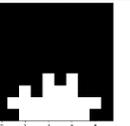<br>326.27 \| 326.14 | 249 | 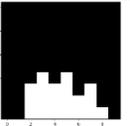<br>326.48 \| 326.22 |
| 5 | - | - | 1125 | 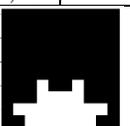<br>326.13 \| 326.06 | 560 | 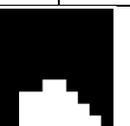<br>326.23 \| 326.11 | 280 | 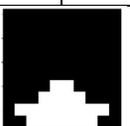<br>326.40 \| 326.09 |
| 6 | - | - | - | - | 622 | 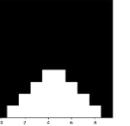<br>326.22 \| 326.06 | 311 | 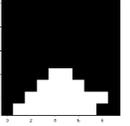<br>326.39 \| 326.08 |
| 7 | - | - | - | - | 684 | 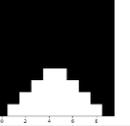<br>326.19 \| 326.06 | 342 | 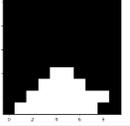<br>326.31 \| 326.08 |
| 8 | - | - | - | - | - | - | 373 | 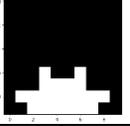<br>326.17 \| 326.06 |
| 9 | - | - | - | - | - | - | 404 | 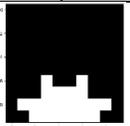<br>326.15 \| 326.06 |

As illustrated in Table 3, all four cases converge and optimal structures with maximum temperature of 326.06K have been found in all cases. The case with the largest $N_0$ converges fastest but needs the largest total number of training data. For cases with small $N_0$, although the total number of training data is small, the convergence is very slow, which implies that it needs more GAN and CNN training. In addition, the convergence may not be monotonic due to the small size of the training data. We conduct two runs of the same design problem with the same $N_0$. Due to random sampling, training samples are different in these two runs. Table 4 and Table 5 list the maximum temperature of the best design in each design iteration obtained from the two runs for cases with $N_0 = 250$ and $N_0 = 125$ respectively. The number of design iterations varies in different run, depending on the training data. In some cases, the maximum temperature of the best design in a later design iteration can be larger than that of the previous iteration. Although it eventually converges, the process is nevertheless much longer. A remedy is to re-train the GAN if the best design identified by the ThemalNet is not better than the best design in the training data of GAN. The rationale for this remedy is that currently there is no standard and consistent criterion to judge the quality of the binary images generate by the GAN other than visual inspection. Re-training the GAN may help the GAN to learn the distribution of the training data better.

Table 4: Maximum temperature of the best design in each design iteration for the case with $N_0 = 250$.

| Different trials | 1st round of inverse design | 3rd round of inverse design | 4th round of inverse design | 5th round of inverse design | 6th round of inverse design | 7th round of inverse design | 8th round of inverse design | 9th round of inverse design |
|---|---|---|---|---|---|---|---|---|
| Trial 1 | 326.20 | 326.15 | 326.09 | 326.08 | 326.06 | - | - | - |
| Trial 2 | 326.47 | 326.34 | 326.24 | 326.19 | 326.15 | 326.08 | 326.09 | 326.06 |

Table 5: Maximum temperature of the best design in each design iteration for the case with $N_0 = 125$

| Different trials | 1st round of inverse design | 2nd round of inverse design | 3rd round of inverse design | 4th round of inverse design | 5th round of inverse design | 6th round of inverse design |
|---|---|---|---|---|---|---|
| Trial 1 | 326.71 | 326.56 | 326.44 | 326.26 | 326.11 | 326.08 |
| Trial 2 | 326.85 | 326.49 | 326.37 | 326.32 | 326.20 | 326.18 |

| Different trials | 7th round of inverse design | 8th round of inverse design | 9th round of inverse design | 10th round of inverse design | 11th round of inverse design | 12th round of inverse design | 13th round of inverse design |
|---|---|---|---|---|---|---|---|
| Trial 1 | 326.06 | - | - | - | - | - | - |
| Trial 2 | 326.11 | 326.09 | 326.09 | 326.09 | 326.14 | 326.08 | 326.06 |

*Computational cost*

The computational cost of the proposed method mainly consists of two parts: time spent on the training of two neural networks and the time spent on generating the training data for the CNN.    As the problem size increases, that is, the number of elements increases, the computational time of the FEM simulations scales linearly with the number of elements. The training time of the networks increases as well, but it does not scale linearly. For an example, the training time for the GAN on 2000 images with 10x10 pixels takes 3.63 minutes. It takes 4.71 minutes on 2000 images with 20x20 pixel. The portion of the FEM simulations increases with the increased problem size, and hence more saving will result.

Table *6* lists the time spent on the training of the GAN, the CNN, and generating the training data (FEM simulations) in each design iteration for the case with $N_0 = 250$. The total time is 118.01 min of which 62.54min spent on the training of the networks and 55.47min spent on generating training data. Here the computational time is evaluated on a computer with 2 x Intel Xeon E5-2670 v3 and 2 x Nvidia Tesla K80. The total time depends on the initial training data $N_0$. For cases with a large $N_0$ and thus a fast convergence rate, the time for network training is less, but the number of training data, thus the FEM simulations, needed is large. Compared with the approach without the CNN, that is, using FEM for temperature evaluation, the total computational time for FEM evaluation alone is 2675min, much larger than 95.47min spent on the training and generating data for the CNN.

As the problem size increases, that is, the number of elements increases, the computational time of the FEM simulations scales linearly with the number of elements. The training time of the networks increases as well, but it does not scale linearly. For an example, the training time for the GAN on 2000 images with 10x10 pixels takes 3.63 minutes. It takes 4.71 minutes on 2000 images with 20x20 pixel. The portion of the FEM simulations increases with the increased problem size, and hence more saving will result.

Table 6: Computational cost for the case with $N_0 = 250$, FEM: time for training data generation, GAN/CNN: training time of GAN/CNN

| Computational time (minutes) | FEM | GAN | CNN |
|---|---|---|---|
| Pre-training for data screening | 22.29 | - | 15.26 |
| 1st iteration | 5.53 | 3.63 | 4.25 |
| 2nd iteration | 5.53 | 3.63 | 4.25 |
| 3rd iteration | 5.53 | 3.63 | 4.25 |
| 4th iteration | 5.53 | 3.63 | 4.25 |
| 5th iteration | 5.53 | 3.63 | 4.25 |
| 6th iteration | 5.53 | 3.63 | 4.25 |
| Total time for each task | 55.47 | 21.78 | 40.76 |
| Total time | 118.01 | | |

## 4.4 Temperature-constrained heat source layout design

The second design problem solved in this work is the temperature-constrained heat source layout design. In this problem, not only the maximum temperature in the entire design domain is to be minimized, the temperature at a certain point should also be higher than a given value, as formulated in Eq. (4).

$$\underset{X}{\text{minimize}}\ T_{max}(X)$$
$$s.t.\quad \Gamma_i \in X, \forall i = 1,2 \dots N \qquad (4)$$

$$T_{point} \geq T_{pmin}$$

where $X$ represents the layout of heat sources, $T_{max}$ is the maximum value of the temperature field in the design domain, $T_{point}$ is the temperature of a given point in the layout, $T_{pmin}$ is the minimum allowed temperature at that point and $\Gamma_i$ denotes the layout area covered by the $i$-th heat source, $N = 20$ is the total number of heat sources. This inverse problem can be reformulated using the KKT approach [46] as:

$$\underset{X}{\text{minimize}} \; f = T_{max}(X) + \lambda(T_{pmin} - T_{point}(X)) \quad (5)$$

where $f$ is the objective function; $\lambda = 0$ if $T_{pmin} - T_{point} \leq 0$ and $\lambda$ is a very large positive value if $T_{pmin} - T_{point} > 0$. In this work, $T_{pmin} = 335K$, $\lambda$ is set to be 10 initially and increases by 10 in each design iteration to reinforce the temperature constraint gradually.

The design domain and the boundary conditions are the same as that of the first design problem presented in Section 4.3. The given point is located at (0.1m, 0.05625m), around the middle of the right side of the square design domain. The origin is set at the left-bottom vertex of the design domain. The parameters of the genetic algorithm used in the optimization process are the same as those listed in Section 4.3.

*Design procedure and results*

The design procedure is the same as that presented in Section 4.3. In the first design iteration, 20k ($M_0 = 20000$) heat source layouts are generated using the random and evolving sampling method. Among them, 125 layouts ($N_0 = 125$) are randomly selected with 75 from random sampling method and 50 from evolving sampling and their temperature fields are calculated using the FEM. These layouts and their temperature fields serves as the initial training data for the ThermalNet. The trained ThermalNet is then used to screen the 20k layouts and the top 2000 layouts, that is, those with the low objective function values, are selected. Layout perturbation is then performed on each selected layout 10 times to produce 20000 new layouts. These new layouts together with the original 2000 layouts go through another screening and the top 2000 ($M_1 = 2000$) are selected and used to train the GAN. Among the 2000 layouts, 31 layouts are randomly selected ($K = 31$) and are added into the initial 125 layouts ($N_1 = N_0 + K = 156$) to form an expanded training set for the ThermalNet. The re-trained ThermalNet and the GAN are then connected to form the design net and the GA is used to obtain design solutions.

In the subsequent design iterations, the design solutions ($Q = 250$) obtained from the previous design iteration are used as the seeds and layout perturbation is performed 20 times on each layout. The newly generated layouts, the design solutions obtained from the previous iteration and the training layouts for the GAN in the previous iteration ($M_{i-1} = 2000$) are merged into one large data set and the ThermalNet screening is performed on this set to select the top 2000 layouts. These 2000 layouts are used as the training data for the GAN in the current design iteration ($M_i = 2000$). For the training of the ThermalNet, again $K$ layouts ($K = 31$) are randomly selected from the current training data for the GAN ($M_i = 2000$) and added to the previous training data ($N_{i-1}$) for the ThermalNet to form the training set in the current iteration ($N_i = N_{i-1} + K$). The trained ThermalNet and GAN are connected to form a new design network and the inverse design is conducted using GA.

Design results together with the total number of training data for the ThermalNet at each design iteration are listed in Table 7. The design process converges within four iterations and the total number of ThermalNet training data used is only 249. The best design obtained from the adaptive learning and optimization approach has the same $T_{max}$ and $T_{point}$, and both are equal to the given $T_{pmin} = 335K$. This fact indicates that the optimal solution of the design problem has been obtained. Similar to what has been done in the first design problem, the influence of the initial training data size for the ThermalNet on the performance of the method in this design problem is investigated. Four design cases with different numbers of the initial training data, ranging from 500 to 62, are conducted. The observed trends are similar to those in the first design problem. All four cases converge and optimal solutions with $T_{max} = T_{point} = T_{pmin} = 335K$ are achieved. Cases with a large set of the initial training data converge fast but need a large number of total training data. Those with a small set of initial training data converge slower, but the total number of training data needed is small as well.

Table 7: Design results at each design iteration: the initial training set for the ThermalNet is 125 and 31 data points are added in each design iteration.

|  | 1st round of inverse design | 2nd round of inverse design | 3rd round of inverse design | 4th round of inverse design |
|---|---|---|---|---|
| Total number of training data | 156 | 187 | 218 | 249 |
| The best design | 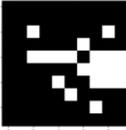 | 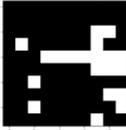 | 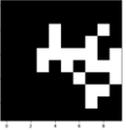 | 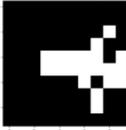 |
| $T_{max}^{NN}$ predicted by ThermalNet/K | 335.34 | 335.28 | 335.12 | 335.15 |
| $T_{point}^{NN}$ predicted by ThermalNet/K | 335.04 | 335.04 | 335.06 | 335.06 |
| $T_{max}^{FEM}$ validated by FEM/K | 335.06 | 335.02 | 335.01 | 335.00 |
| $T_{point}^{FEM}$ validated by FEM/K | 335.06 | 335.00 | 335.00 | 335.00 |

*Performance comparison with other layout design approaches*

The performance of the proposed neural network-based adaptive design approach is compared with the neural network-based method (GAN+CNN+GA) without adaptive learning. For a fair comparison, the total amount of training data used is roughly the same. As shown in Table 8, the maximum temperature obtained from the approach without adaptive learning is 335.17K, which is larger than the temperature at the given point 335.06K. In addition, even though the temperature constraint is satisfied, the design is clearly not optimal, which illustrates the effectiveness of the adaptive learning strategy.

The proposed adaptive design approach also shows advantages as compared with the method listed in Ref. [4] in which a neural network is constructed to predict the temperature field. In their work, 55000 data points are used to train the network, and the best design obtained has a maximum temperature of 335.07K, which is larger than the optimal solution 335K. With the proposed approach, the optimal solution can be obtained using only 249 training data, much less than that used in Ref. [4].

Table 8: Design results obtained from different layout design methods (FPN-NSLO: Feature-pyramid-network and neighborhood search-based layout design [4])

|  | Number of FEM calculations | $T_{max}$ of the best design/K | $T_{point}$ of the best design/K |
|---|---|---|---|
| FPN_NSLO | 55000 | 335.07 | 335.07 |
| GAN+ CNN+ GA without adaptive learning | 249 | 335.17 | 335.06 |
| GAN+ CNN+ GA with adaptive learning | 156 | 335.06 | 335.06 |
|  | 249 | 335.00 | 335.00 |

## 5. Summary and future work

In this work, an adaptive artificial neural network-based generative design approach has been proposed and developed for layout design. This method combines the state-of-the-art generative adversarial network and the convolutional neural network with the genetic algorithm and uses an adaptive learning and optimization strategy to effectively explore the design space and find desired solutions. As such, the number of high-dimensional calculations for the evaluation of the objective function is greatly reduced. The proposed design approach has been applied to solve two heat source layout design problems in which 20 square-shaped heat sources are to be distributed within the design domain. The design objective of both problems is to minimize the maximum temperature of the entire domain. The second problem has an additional temperature constrain at a given point. In both problems, optimal designs have been obtained. Compared with several existing approaches including both conventional and neural network-based methods, the proposed approach has the best performance in the terms of generality, accuracy and efficiency.

Layout designs are encountered in a variety of different fields as mentioned in the introduction. The proposed design approach should be applicable for a general set of layout designs. Future work will include the applications of the proposed approach to problems with a large number of heat sources of different shapes and the design of composite materials with unique properties.

Acknowledgement

This work is supported by the Hong Kong Research Council under Competitive Earmarked Research Grants No. 16212318 and No. 16206320.

References

[1] P Lissaman, CA Shollenberger. Formation flight of birds, Science. 168 (1970) 1003-1005.

[2] P Seiler, A Pant, JK Hedrick. A systems interpretation for observations of bird V-formations, J.Theor.Biol. 221 (2003) 279-287.

[3] C Chen, GX Gu. Generative deep neural networks for inverse materials design using backpropagation and active learning, Adv. Sci. 7 (2020) 1902607.

[4] X Chen, X Chen, W Zhou, J Zhang, W Yao. The heat source layout optimization using deep learning surrogate modeling, Struct. Multidiscip. Optim. 62 (2020) 3127–3148.

[5] W Tian, Z Mao, F Zhao, Z Zhao. Layout optimization of two autonomous underwater vehicles for drag reduction with a combined CFD and neural network method, Complexity. 2017 (2017).

[6] GX Gu, C Chen, MJ Buehler. De novo composite design based on machine learning algorithm, Extreme Mech. Lett. 18 (2018) 19-28.

[7] C Qian, W Ye. Accelerating gradient-based topology optimization design with dual-model artificial neural networks, Struct. Multidiscip. Optim. (2020) pp1-21.

[8] H Chi, Y Zhang, TLE Tang, L Mirabella, L Dalloro, L Song, et al. Universal machine learning for topology optimization, Comput.Methods Appl.Mech.Eng. 375 (2020) 112739.

[9] O Hennigh. Automated design using neural networks and gradient descent, arXiv preprint arXiv:1710.10352. (2017).

[10] J Peurifoy, Y Shen, L Jing, Y Yang, F Cano-Renteria, BG DeLacy, et al. Nanophotonic particle simulation and inverse design using artificial neural networks, Sci. Adv. 4 (2018) eaar4206.

[11] H Zhou, D Alexander, K Lange. A quasi-Newton acceleration for high-dimensional optimization algorithms, Statistics and computing. 21 (2011) 261-273.

[12] MN Omidvar, X Li, K Tang. Designing benchmark problems for large-scale continuous optimization, Inf.Sci. 316 (2015) 419-436.

[13] Y Zhang, W Ye. Deep learning–based inverse method for layout design, Struct. Multidiscip. Optim. 60 (2019) 527-536.

[14] RK Tan, NL Zhang, W Ye. A deep learning–based method for the design of microstructural materials, Struct. Multidiscip. Optim. 61 (2019) 1417–1438.

[15] X Li, S Ning, Z Liu, Z Yan, C Luo, Z Zhuang. Designing phononic crystal with anticipated band gap through a deep learning based data-driven method, Comput.Methods Appl.Mech.Eng. 361 (2020) 112737.

[16] L Wang, Y Chan, F Ahmed, Z Liu, P Zhu, W Chen. Deep generative modeling for mechanistic-based learning and design of metamaterial systems, Comput.Methods Appl.Mech.Eng. 372 (2020) 113377.

[17] W Wang, X Zhang, C Xin, Z Rao. An experimental study on thermal management of lithium ion battery packs using an improved passive method, Appl.Therm.Eng. 134 (2018) 163-170.

[18] L Cai, W Chen, G Du, X Zhang, X Liu. Layout design correlated with self-heating effect in stacked nanosheet transistors, IEEE Trans.Electron Devices. 65 (2018) 2647-2653.

[19] X Zhu, C Zhao, X Wang, Y Zhou, P Hu, Z Ma. Temperature-constrained topology optimization of thermo-mechanical coupled problems, ENG OPTIMIZ. 51 (2019) 1687-1709.

[20] J Sun, J Zhang, X Zhang, W Zhou. A Deep Learning-based Method for Heat Source Layout Inverse Design, IEEE Access. (2020).

[21] DW Hengeveld, JE Braun, EA Groll, AD Williams. Optimal placement of electronic components to minimize heat flux nonuniformities, J.Spacecraft Rockets. 48 (2011) 556-563.

[22] V Lau, FLd Sousa, RL Galski, EM Rocco, JC Becceneri, WAd Santos, et al. A multidisciplinary design optimization tool for spacecraft equipment layout conception, J. Aerosp. Technol. Manag. 6 (2014) 431-446.

[23] DF Hanks, Z Lu, J Sircar, TR Salamon, DS Antao, KR Bagnall, et al. Nanoporous membrane device for ultra high heat flux thermal management, Microsyst. Nanoeng. 4 (2018) 1-10.

[24] A Mazloomi, F Sharifi, MR Salimpour, A Moosavi. Optimization of highly conductive insert architecture for cooling a rectangular chip, Int.Commun.Heat Mass Transfer. 39 (2012) 1265-1271.

[25] L Chen, Y Sun, J Lin, X Du, G Wei, S He, et al. Modeling and analysis of synergistic effect in thermal conductivity enhancement of polymer composites with hybrid filler, Int.J.Heat Mass Transfer. 81 (2015) 457-464.

[26] T Dbouk. A review about the engineering design of optimal heat transfer systems using topology optimization, Appl.Therm.Eng. 112 (2017) 841-854.

[27] K Chen, S Wang, M Song. Optimization of heat source distribution for two-dimensional heat conduction using bionic method, Int.J.Heat Mass Transfer. 93 (2016) 108-117.

[28] K Chen, J Xing, S Wang, M Song. Heat source layout optimization in two-dimensional heat conduction using simulated annealing method, Int.J.Heat Mass Transfer. 108 (2017) 210-219.

[29] K Chen, S Wang, M Song. Temperature-gradient-aware bionic optimization method for heat source distribution in heat conduction, Int.J.Heat Mass Transfer. 100 (2016) 737-746.

[30] Y Aslan, J Puskely, A Yarovoy. Heat source layout optimization for two-dimensional heat conduction using iterative reweighted L1-norm convex minimization, Int.J.Heat Mass Transfer. 122 (2018) 432-441.

[31] RR Madadi, C Balaji. Optimization of the location of multiple discrete heat sources in a ventilated cavity using artificial neural networks and micro genetic algorithm, Int.J.Heat Mass Transfer. 51 (2008) 2299-2312.

[32] T Sudhakar, C Balaji, SP Venkateshan. Optimal configuration of discrete heat sources in a vertical duct under conjugate mixed convection using artificial neural networks, Int. J. Therm. Sci. 48 (2009) 881-890.

[33] YH Lee, AC Marvin, SJ Porter. Genetic algorithm using real parameters for array antenna design optimisation, High Frequency Postgraduate Student Colloquium. (1999) pp8-13.

[34] R Chelouah, P Siarry. A continuous genetic algorithm designed for the global optimization of multimodal functions, J.Heuristics. 6 (2000) 191-213.

[35] K Deb, A Pratap, S Agarwal, T Meyarivan. A fast and elitist multiobjective genetic algorithm: NSGA-II, IEEE Trans. Evol. Comput. 6 (2002) 182-197.

[36] I Goodfellow, J Pouget-Abadie, M Mirza, B Xu, D Warde-Farley, S Ozair, et al. Generative adversarial nets, Adv Neural Inf Process Syst. 27 (2014) 2672-2680.

[37] K Cheng, R Tahir, LK Eric, M Li. An analysis of generative adversarial networks and variants for image synthesis on MNIST dataset, Multimedia Tools Appl. (2020) pp1-28.

[38] Y Jin, J Zhang, M Li, Y Tian, H Zhu, Z Fang. Towards the automatic anime characters creation with generative adversarial networks, arXiv preprint arXiv:1708.05509. (2017).

[39] F Wen, J Jiang, JA Fan. Progressive-Growing of Generative Adversarial Networks for Metasurface Optimization, arXiv preprint arXiv:1911.13029. (2019).

[40] S Rawat, MH Shen. A novel topology optimization approach using conditional deep learning, arXiv preprint arXiv:1901.04859. (2019).

[41] Y Zhang, A Chen, B Peng, X Zhou, D Wang. A deep Convolutional Neural Network for topology optimization with strong generalization ability, arXiv preprint arXiv:1901.07761. (2019).

[42] T Hsu, WK Epting, H Kim, HW Abernathy, GA Hackett, AD Rollett, et al. Microstructure Generation via Generative Adversarial Network for Heterogeneous, Topologically Complex 3D Materials, JOM. 73 (2021) 90-102.

[43] M Mirza, S Osindero. Conditional generative adversarial nets, arXiv preprint arXiv:1411.1784. (2014).

[44] J Hu, L Shen, G Sun. Squeeze-and-excitation networks, Proc. IEEE Comput. Soc. Conf. Comput. Vis. Pattern Recognit. (2018) pp7132-7141.

[45] DP Kingma, J Ba. Adam: A method for stochastic optimization, arXiv preprint arXiv:1412.6980. (2014).

[46] HW Kuhn, AW Tucker. Nonlinear programming, in (j. neyman, ed.) proceedings of the second berkeley symposium on mathematical statistics and probability, University of California Press, Berkeley. (1951).